\documentclass[twocolumn,10pt]{article}
\ifdefined\pdfsuppressptexinfo\pdfsuppressptexinfo=-1\fi
\ifdefined\pdfinfoomitdate\pdfinfoomitdate=1\fi
\ifdefined\pdftrailerid\pdftrailerid{}\fi
\usepackage[letterpaper,margin=0.9in,columnsep=0.28in]{geometry}
\usepackage{times}
\usepackage{graphicx,amsmath,microtype}
\usepackage{natbib}
\usepackage[hidelinks]{hyperref}
\usepackage{xurl}
\newcommand{\sect}{\S}
\title{Compiled Agency: Coding Agents as Game AI Researchers\,---\,from a Roguelike to StarCraft~II and Civilization}
\date{}

\begin{document}

\twocolumn[{%
\begin{center}
{\LARGE\bfseries Compiled Agency: Coding Agents as Game AI Researchers\,---\,from a Roguelike to StarCraft~II and Civilization\par}
\vspace{1.1em}
{\large Haonan Huang$^{1}$ \quad Joey Xiao$^{2}$\par}
\vspace{0.35em}
{$^{1}$Princeton University \qquad $^{2}$New York University\par}
\vspace{1.3em}
\begin{minipage}{0.88\textwidth}
\small\textbf{Abstract.}\enspace Coding agents are increasingly capable of sustained engineering and empirical research. Greater autonomy makes their research decisions themselves a target for evaluation: what to investigate, which experiments to run, and when to stop. Games provide a controlled, interactive setting with objectively measurable outcomes. We introduce Gauntlet, a develop--freeze--evaluate protocol for studying these decisions as agents build standalone game-playing programs. Starting from a game description, a raw observation/action interface, and an empty policy file, an off-the-shelf agent develops a controller from bare interaction in one autonomous session. We call the capability under study \emph{compiled agency}: the shipped program plays with zero model calls. Every intermediate version is frozen for later scoring on held-out instances, connecting the agent's research process to independently measured performance. The capability is real and advancing: environment access adds 10 to 78 percentage points of held-out success over construction-only controls, and progress across model generations comes in steps, with tiers that defeat one generation entirely falling to the next. The protocol yields StarCraft~II controllers that defeat every fair built-in AI, and in the newest generation the strongest cheating tier as well, and Civilization controllers that win complete games by conquest. Replaying more than 5{,}000 frozen versions exposes the research behind the programs: gains that plateau early; rigorous local investigation beside sparse validation of what actually ships; and stops that follow a race between the agent's own evidence and a model-specific transcript budget it was never given. When validation panels are refreshed mid-session, so that only the evidence changes, shipped success rises 12.4 points in nine of nine completed pairs of twelve initiated. The experiments an agent designs are part of the capability it delivers; Gauntlet makes them measurable and improvable---a step toward agents whose research practice, not just whose code, can be engineered. We release the benchmark, the corpus, and the replay tooling.
\end{minipage}
\vspace{1.6em}
\end{center}
}]

\let\thefootnote\relax\footnotetext{Appendix pointers (A--N) refer to the extended version of this paper.}

\section{Introduction}
\label{sec:intro}

Coding agents now handle increasingly long-horizon work \citep{kwa2025metr}, and increasingly that work is empirical research: probing systems, forming and testing explanations, and deciding when the evidence is sufficient \citep{lu2024aiscientist,huang2026grounded,huang2026scrutiny}. Assessing this capability requires connecting the agent's investigations to what its work achieves. A final score alone does not reveal how the capability developed or whether the agent's own evidence justified its conclusions. We therefore seek a setting where agents direct their own investigations, outcomes are free of external confounds, and the programs they produce can be evaluated independently throughout development.

Game-controller development offers such a setting. Games provide interactive feedback and objectively measurable outcomes under experimental control, and successive controllers preserve executable consequences of the agent's development decisions. Existing work evaluates models playing games directly \citep{paglieri2025balrog,hu2026lmgame,ying2026gamestore} and synthesizes executable policies within supplied representations and refinement procedures \citep{xu2025portal,deng2024smacr1}. We instead leave the entire process to the agent: an off-the-shelf coding agent constructs the complete controller and organizes its own experiments in a single session of bare interaction. The game panel spans difficulty and scale: at one end a single-file roguelike whose trials cost seconds; at the other, StarCraft~II and Civilization, where local improvements count only once integrated into a winning full-game controller. We call the capability \emph{compiled agency}: the shipped program plays with zero model calls.

\begin{figure*}[t]\centering
\includegraphics[width=5.5in]{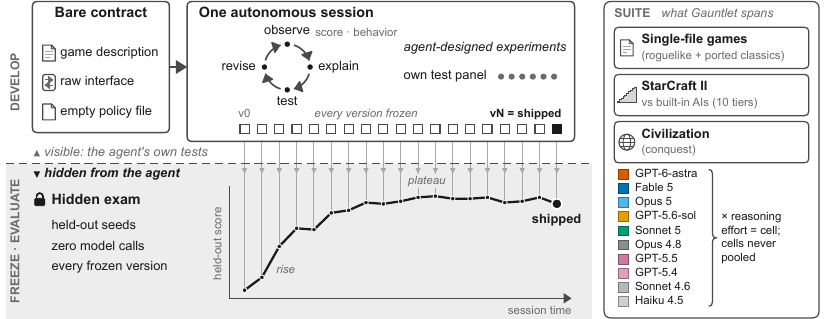}
\caption{\textbf{The Gauntlet protocol.} An agent receives a game description, a raw interface, and an empty policy file, and develops a standalone controller in one autonomous session (left); every version is frozen for replay on hidden held-out instances with zero model calls, at the per-game coverage of Appendix~B (center). Right: environments and models; a cell (model $\times$ effort $\times$ game) is the unit of analysis, never pooled.}
\label{fig:protocol}
\end{figure*}

Gauntlet, the protocol introduced here, connects the development process to independent evaluation. The agent works under a bare contract (\sect\ref{sec:setup}), with no stated budget and no clock shown. Intermediate versions are frozen and replayed afterwards on held-out instances unavailable during development. The trace records what the agent tested and concluded; the replay measures what its programs can do; comparing the two tests whether the agent's evidence supports what it ships (Figure~\ref{fig:protocol}).

Freezing everything turns qualitative impressions into five measurable questions. What does the research produce? A leaderboard with recurring generational steps and collapsing cost per win (\sect\ref{sec:produces}). What does it consist of? Auditable observation--explanation--test--revision chains (\sect\ref{sec:consists}). What do they achieve? Plateauing curves, measurement that tracks the exam beside optimistic final claims, and stale final validation (\sect\ref{sec:achieves}). What ends a session? A race between the agent's evidence and an uninstructed transcript budget (\sect\ref{sec:stopping}). Does changing the evidence change the product? Yes, where headroom and uptake meet (\sect\ref{sec:intervention}).

Our contributions are the protocol and suite, a version-level replayed corpus, the diagnosis above, and one demonstrated remedy. The corpus holds about 5{,}000 scored versions from roughly 140 development sessions, several hundred audited research episodes, and a 1{,}110-session stopping inventory spanning 19 games (scoring coverage per game: Appendix~B). Throughout, ``researcher'' describes behavior, not cognition. Every performance statistic is reported within a cell (one model, one reasoning effort, one game and condition); corpus-wide summaries are labeled as inventories, because pooling across cells repeatedly manufactured conclusions that no single cell supports.

\section{Protocol, corpus, and definitions}
\label{sec:setup}

\textbf{Environments.} The suite spans three families of games, ordered by the cost of a single trial. The single-file family holds a privately written roguelike that has never been published or shared, and fifteen ported single-file games; their engines are deterministic, so the roguelike's 80-seed held-out exam is exact (identical artifacts receive identical per-seed outcomes), scored as clears (R1) and as a speed-weighted utility (R2; Appendix~B). In StarCraft~II the agent writes a Python controller against the raw low-level API \citep{vinyals2017sc2le,blizzard2017s2clientproto}, with no bot framework provided \citep[e.g.,][]{burnysc2pythonsc2}, and is examined on 16 held-out scenarios (unseen maps and seeds) at its development tier. In Civilization it plays full conquest games against built-in AIs through a web-served engine (Freeciv, via CivRealm; \citealp{qi2024civrealm}), with a 16-game held-out exam per condition; a single trial takes minutes to hours.

\textbf{Contract.} Every session starts from the same bare contract: the description, the interface, an empty policy file, and general-purpose tools. Nothing else is provided: no baseline, no starter code, no budget, no clock. The prompt says the exam will use unseen instances but not which. Whatever is in the policy file when the session ends is what gets examined.

\textbf{Cells, harnesses, controls.} A cell is model $\times$ reasoning effort $\times$ game. We call the highest-effort cells of the current model generations the \emph{frontier cells}; on the roguelike these have $N{=}6$ sessions each (other cells' $N$ stated where used). Each model runs under its vendor's command-line agent (Claude Code for the Claude models, codex for the GPT models) at a fixed effort per cell, the highest available unless stated; the harness's own context management, including compaction, is part of the environment the agent experiences (\sect\ref{sec:stopping}). Control and instrument runs (one-shot, spec-only, uncapped, and the forked branches of \sect\ref{sec:intervention}) are analyzed separately and never enter cell statistics.

\textbf{Replay.} Intermediate versions are reconstructed from the session traces and checked: the reconstruction's final state must byte-match the shipped artifact, and every snapshot the agent left in its workspace must byte-match the reconstruction at that moment; all reconstructions used here pass both (one session with whitespace-only differences; Appendix~B). StarCraft~II replays also reproduce the recorded exam of the shipped build in 93\% of games (252/272); the shortfall is cross-build engine nondeterminism (byte-identical builds agree 16/16).

\textbf{Definitions.} A \emph{research episode} is a trace segment with an observation, a stated explanation, an executed discriminating test, and a decision (\sect\ref{sec:consists}). A session is \emph{self-terminated} when the agent ends it; wall-capped, quota-killed, and operator-stopped sessions are excluded from the natural-stop fits, right-censored in the registered hazard analysis, and retained, labeled, in the corpus. Recovery of the agent's own evaluations from traces is in Appendix~J.

\textbf{Disclosures.} Despite instructions discouraging it, a small fraction of canonical roguelike trials (30 of 292) read the minified environment source \citep[cf.][]{zhong2025impossiblebench}; those sessions gain no statistical advantage on the exam (Appendix~M).

\section{What the research produces}
\label{sec:produces}

\begin{figure*}[t]\centering
\includegraphics[width=5.5in]{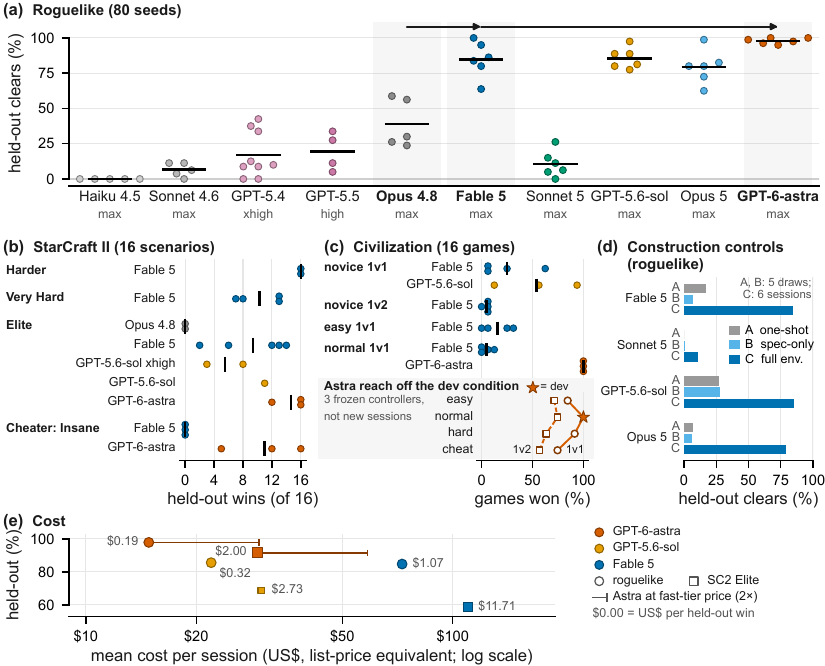}
\caption{\textbf{Shipped controllers across three games.} \textbf{(a)} Roguelike: one dot per shipped controller, per cell at each model's headline effort (GPT-5.5: high; Appendix~D); ticks are cell means. \textbf{(b)} StarCraft~II: held-out wins of 16 per session (recorded exams), by development tier and model. \textbf{(c)} Civilization: conquest wins of 16 per session, by condition; inset: Astra's normal-1v1 controllers across seven other conditions. \textbf{(d)} Construction controls: one-shot (A), spec-only (B), full environment (C). \textbf{(e)} Cell mean vs list-price dollars per session; whisker: Astra's unknown tier up to $2\times$.}
\label{fig:leaderboard}
\end{figure*}

\textbf{The protocol produces programs worth ranking} (Figure~\ref{fig:leaderboard}). On the roguelike, cell means at each model's headline effort run from 0\% (Haiku~4.5) to 97.9\% (GPT-6-astra, hereafter Astra), and the widest separations in Figure~\ref{fig:leaderboard}a are generational. Within each vendor lineage the matched-cell step is large: Opus~4.8 to Fable~5 goes from 39.0\% to 84.8\%, and GPT-5.6-sol to Astra from 85.6\% to 97.9\% (470/480 seeds), the spread collapsing to a four-clear band reached in 0.5--1.1 hours per session against Fable~5's 1.5--7.5 (Appendix~F). Construction controls locate the capability: one-shot generation averages at most 27\% in its best cell, spec-only construction changes little, and full environment access adds 10 to 78 points depending on the model (Figure~\ref{fig:leaderboard}d); the margin is earned in the environment.

StarCraft~II tells the same story against a harder wall. In Figure~\ref{fig:leaderboard}b solved tiers stack at the top (Fable~5 clears Harder perfectly, 64\% at Very Hard) while tiers at the edge of a model's reach scatter. At Elite, where Opus~4.8 scored zero, Fable averages 59\%, and Astra's three sessions score 12, 16, and 16 of 16, the first perfect exams at that tier. At Cheater:Insane, the highest difficulty, every Fable session shipped a controller that won nothing; Astra ships 16, 5, and 12 of 16, a tier that defeated one generation entirely falling to the next. Session-to-session variance, which we call the \emph{session lottery}, persists exactly where headroom remains. Civilization repeats the pattern at full-game scale (Figure~\ref{fig:leaderboard}c): Fable wins complete conquest games at every condition it attempted, with low, lottery-like counts (normal 1v1 mean 4.7\%); GPT-5.6-sol's three novice-condition sessions score 9, 15, and 2; and Astra's three sessions at normal 1v1, the hardest condition attempted, score 16, 16, and 16 of 16 (watchdog reruns: Appendix~C).

A frozen controller can also be examined away from its development difficulty, turning each shipped program into a reach profile: Astra's best Cheater:Insane controller (its 16/16 session) sweeps all ten StarCraft~II difficulties (59 of 60) while its Elite controllers manage 0--1 of 6 at Insane, and its normal-1v1 Civilization controllers win 74\% of decided games across the seven other conditions (221/300; 36 of 336 unresolved) against 100\% at home, losing ground mainly to a second opponent, a deficit that survives any resolution of the unresolved games (Appendix~C). In StarCraft~II, reach tracks the development tier.

Final performance is one axis of the leaderboard; the cost of reaching it is the other. In Figure~\ref{fig:leaderboard}e Astra sits far left of Fable~5 at equal or better quality: \$14.85 against \$72.80 per roguelike session ($4.9\times$), \$0.19 against \$1.07 per held-out clear ($5.6\times$). List prices per token are identical, so the saving is volume; Astra consumes 17--20\% of its comparators' input tokens. From sol to Astra, cost per clear falls from \$0.32 to \$0.19 while quality rises (Appendix~E).

\section{What the research consists of}
\label{sec:consists}

\textbf{The sessions contain real, countable investigation.} We audited research episodes from the agents' own messages and commands (hidden reasoning is never used); an episode requires an executed, discriminating test, not a narration. A keyword-guided pass over 39 sessions labeled 204 episodes; complete re-reads of two Civilization sessions counted 31 and 127 where that pass had found 9 and 7, about one strict episode per policy version the agent played. The corpus therefore holds several hundred episodes, five or so in an hours-long session and over a hundred in a long Civilization one (Appendix~F). In the re-read sessions, looser tune-and-test loops add only 10--20\%: those agents state a mechanism and run a check for most changes. About a quarter of episodes end with the agent revising or rejecting its own explanation (26\% in both full re-reads), the signature of investigation rather than storytelling; a second reader blinded to session and outcome reproduced nine of ten sampled status labels. In the audited roguelike sessions, which every model plays, frontier models build instruments while weaker models tune constants and re-run; the big games admit only the strongest models (Appendix~F).

At its best the work is recognizable research, drawing on its two raw materials, outcome numbers and observed behavior; one audited session shows each used well. In StarCraft~II the alarm was a number: a win that consumed 46{,}824 of 48{,}000 allowed loops, nearly a timeout. Rather than tune parameters, the agent re-ran the scenario with instrumentation and found the cause: its army was ``attacking'' with an empty memory of enemy structures, wandering instead of hunting. It added a map-covering search lattice with parallel hunt squads and won the retest in 67\% fewer loops. The search mode shipped, and the controller scored 14/16 on the hidden exam.

The Civilization exemplar starts from behavior instead of a number, and it ends the other way: a sound local investigation attached to a failing program. Five newly built settlers had each died on the AI turn after completion. The agent blamed enemy attacks and built escorts, twice, and both fixes failed in replay. A client log line then surfaced the real cause: the settlers were being \emph{disbanded} for lack of food upkeep. The agent reversed itself explicitly (``this was not combat''), changed the build rule to require a food surplus, and verified in a fresh replay of the same seed that a new settler survived. As local investigation, little is missing. The shipped controller scored 2/16. Local rigor does not bound the exam. The starkest case is a GPT-5.6-sol session at easy 1v1 that spent roughly 190 hours and 50{,}000 tool calls, played some 300 policy versions and wrote 1{,}200 more candidate files, ran until stopped externally, and shipped a policy scoring 0/16 while its two cell-mates scored 12 and 16. Only the held-out exam tells such sessions apart.

\section{What the investigations achieve}
\label{sec:achieves}

\begin{figure*}[t]\centering
\includegraphics[width=5.5in]{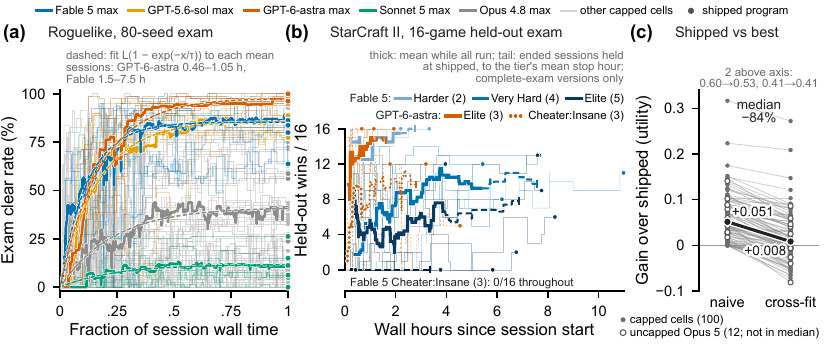}
\caption{\textbf{Measured progress.} \textbf{(a)} Roguelike held-out score against session wall fraction, per cell (100 capped sessions, 4{,}114 versions, plus Astra's 6 sessions, 99 versions); dashed: the two-parameter fits; dots mark shipped programs. \textbf{(b)} StarCraft~II held-out wins against wall hours (complete-exam versions only): Fable~5 by tier (thin: sessions; thick: tier means; $n$ counts sessions) with Astra's Elite and Cheater:Insane sessions drawn separately: a plateau where the tier is within reach, none at its edge. \textbf{(c)} Shipped versus best checkpoint, in exam utility: the apparent gap collapses under cross-fitting.}
\label{fig:process}
\end{figure*}

\textbf{Curves are a property of the cell.} Replaying every roguelike version measures every policy the agent wrote, independently of its own testing; trajectories are jagged mostly because policies really change between versions, so sessions are described as draws around a cell curve. Figure~\ref{fig:process}a shows the shape directly: the thick curves, each a cell average, rise steeply in the first quarter of a session and flatten, while the thin individual trajectories scatter around them. A simple two-parameter saturating fit captures the averages (dashed in Figure~\ref{fig:process}a; alternatives fit no better, Appendix~H), its level recovering the drawn cell mean within 1.2 points where cells plateau. The plateaus are long: each frontier cell reaches nine tenths of its final level within the first quarter to two fifths of the session and holds it for the rest, and Astra compresses the whole shape (75/80 by its fifth to eighth version). StarCraft~II shows the same rise where the tier is within reach: in Figure~\ref{fig:process}b the Very Hard mean (four sessions) climbs from 2 to 8 wins in three hours and ends near 11 of 16; Astra's Elite mean reaches 14.3 within the hour. At the edge of reach there is no plateau: Fable's Elite mean dips and recovers, one session falling from 12 wins to 0 before ending at 13, while Astra's Cheater:Insane sessions climb to a shipped mean of 11 of 16 where every Fable session stays at zero.

\textbf{Late progress recycles old ground.} A step's gains are the exam seeds that flip to cleared between consecutive versions; its losses, the seeds that flip to failed. Early in a session, 13--90\% of gained seeds (by frontier cell) are first-time clears; by the final third that share is essentially zero in all 21 cells, so late gains recover seeds already won and lost. The per-seed loss hazard falls rather than rises: the plateau is damped churn over a shrinking contested pool, not growing interference; instrument sessions, run far past the normal horizon, decay to exhaustion (Appendix~I).

\textbf{Shipping loses little to checkpoint selection.} The worry that a better version was left behind is mostly a same-seed artifact: picking the ``best'' checkpoint on the seeds that then score it overstates the gap. Figure~\ref{fig:process}c estimates what a 40-seed selector could actually recover, choosing on one seed half and scoring on the other: the median gain over shipping falls from $+0.05$ naive to $+0.01$ cross-fit in exam utility, with a real tail (36 of 100 sessions gain $+0.02$ or more; this measures that selector, not an oracle). On the 16-game StarCraft~II exam, where ``best earlier'' is itself partly a draw of finite-sample noise, shipped matches or beats it in 5 of 11 Fable and 2 of 3 Astra sessions; exceptions are real (one Astra session held a 16/16 at 0.2 hours and shipped 12/16).

\begin{figure*}[t]\centering
\includegraphics[width=5.5in]{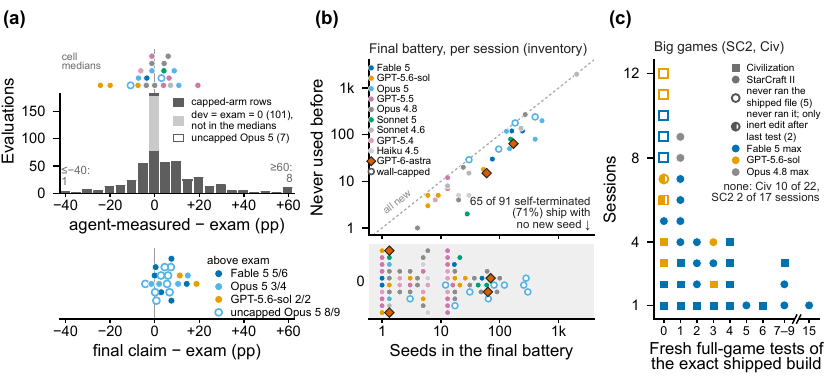}
\caption{\textbf{The agent's evidence.} \textbf{(a)} Agent-measured minus hidden-exam score on the same policy (top) and final claims minus exam (bottom). \textbf{(b)} The final validation battery per session: battery size vs seeds never used before; shaded strip = zero new seeds. \textbf{(c)} Fresh full-length tests of the exact shipped build in the big games; of the 12 with none, 7 never ran the shipped file (Appendix~J).}
\label{fig:evidence}
\end{figure*}

\textbf{Measurement shows no consistent bias; final claims are optimistic.} On the 527 policies agents chose to measure, in-development numbers carry no consistent optimistic bias across cells. The top panel of Figure~\ref{fig:evidence}a is centered on zero, with cell medians on both sides: ten above, ten below, $-24$ to $+19$ points, tracking the hardness of each cell's own test seeds relative to the exam (small panels add scatter). The strip beneath it sits to the right: final self-claims exceed the exam in 10 of 12 capped-cell claims and 8 of 9 uncapped claims (21 claims from 18 sessions), and exceed the cell's own measured median in all six cells with claims. The final battery, the last evaluation of the exact shipped file, is where the evidence is weakest: 65 of 91 self-terminated roguelike sessions ship without a single seed they had never tried (Figure~\ref{fig:evidence}b), and battery size is a model habit (medians 6--138 by model), not a response to uncertainty. Astra is the exception on size (60--171-seed batteries in four sessions) but not on freshness; twice it edited the file after the battery and shipped on a single-seed check.

The big games show the same staleness where evidence is dearest: one full trial costs minutes to hours, and agents ship on thin direct evidence (Figure~\ref{fig:evidence}c). The median number of fresh full-horizon tests of the exact shipped build is 1.5 per session for Fable on Civilization and 0 for GPT-5.6-sol there (5 of 16 and 5 of 6 sessions ran none). On StarCraft~II Fable's median is 2. How much an agent validates does not predict its exam score across cells; which instances it validates on does (\sect\ref{sec:intervention}).

\section{How sessions end}
\label{sec:stopping}

\begin{figure*}[t]\centering
\includegraphics[width=5.5in]{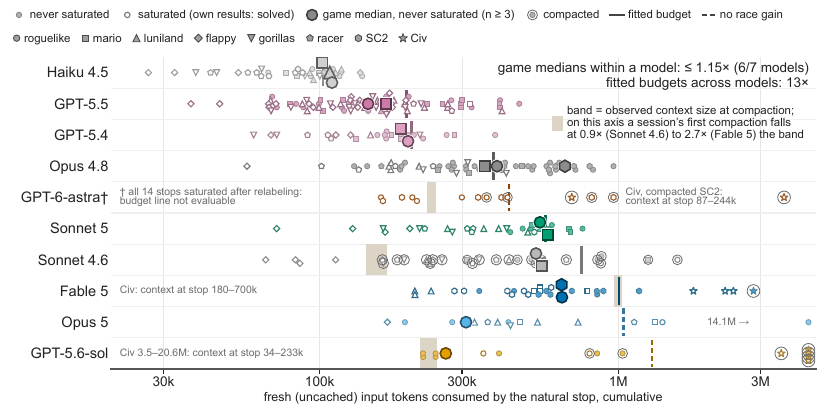}
\caption{\textbf{Where sessions stop.} Natural stops per model across games, in cumulative fresh input tokens: filled = never saturated, hollow = saturated, large = per-game medians; line = fitted transcript budget (dashed: no race gain); band = context at observed compactions (absent: model never compacts); rings = compacted (past-band, ring-free stops = cache re-reads; $\dagger$Astra: all 14 stops saturated, budget not evaluable). Never-saturated medians align across games within a model; budgets span $13\times$ across models; saturated stops fall short by a game-dependent wrap-up.}
\label{fig:stopping}
\end{figure*}

\textbf{Stopping follows a race between evidence and an uninstructed, model-specific scale.} No session is told a budget or a deadline (external wall caps exist but are never announced; clock fragments leak into 95 of 101 audited sessions, and exactly one agent ever reasoned about absolute time, not about stopping). Call a session \emph{saturated} once the agent has seen a result showing the task solved (roguelike: one of its own batteries $\ge$95\% cleared; StarCraft~II: consecutive full-game wins; Appendix~K). Sessions that never saturate (91\% of roguelike sessions, most harder StarCraft~II tiers) stop at a model-specific scale of consumed transcript; we call the fitted scale an \emph{effective transcript budget}, a predictive description rather than an identified internal mechanism. Fitted budgets run from 102k fresh input tokens (Haiku~4.5) to about 1.3M (GPT-5.6-sol), a $13\times$ span (Figure~\ref{fig:stopping}). The regularity is an alignment of medians: within each row of Figure~\ref{fig:stopping} the per-game median markers align (within $1.15\times$ for six of seven multi-game models) while sessions still spread two- to three-fold inside a cell. Model identity sets the scale of stopping; the game barely moves it; nothing pins a single session. Saturated sessions stop earlier, after a wrap-up of continued polishing. That wrap-up is failure-gated, not threshold-timed: sessions almost never end on a failing result and polish longer where the game leaves more to gain (Appendix~K). A race model, stopping at whichever comes first of saturation-plus-wrap-up and the budget, beats the budget alone out of sample in 7 of 10 fitted models (7 of 9 evaluable): the budget carries the never-saturated stops, and the evidence term adds the saturated ones (a registered hazard analysis agrees, and no visible progress signal adds more; Appendix~K).

The budget is a scale of consumed transcript, not of context: cache re-reads bill old context again, so cumulative counts pass triggers that live context never reaches. The triggers are directly observable ($\approx$244k context under codex; 143--168k for Claude Code with older Claude models; Appendix~K), most sessions stop with live context below them, and only 38 of 330 drawn stops ever compact, 24 of them Sonnet~4.6's (Appendix~K). Crossing the trigger splits behavior (rings in Figure~\ref{fig:stopping}): Sonnet~4.6 compacts routinely yet still stops at a stable cumulative scale, while GPT-5.6-sol's Civilization sessions cross it dozens of times and become long-lived, ending at up to sixteen times the budget.

What happens at the stop is a confirmation routine, not a threshold. In the nine models other than Astra, finalization language precedes the stop more often than it precedes other evaluations, and in seven of nine the last evaluation re-runs an unchanged policy, mostly on seeds already used: a fresh measurement of nothing new. Whether a failing final check delays the stop is model-specific (Appendix~K); Astra is the strict case, never once stopping after a failing battery: every failure led to an edit and retest (19/19 final games won). Astra is also the model whose stops the evidence explains outright: with its background-shell batteries recovered from the traces, all 14 of its natural stops are saturated. It solves the task quickly, confirms, and leaves; its budget is unidentifiable: it never runs long enough to reveal one. For every other model the summary is blunt: how long an agent works is predicted mostly by which model it is, and the plateau of Figure~\ref{fig:process} is held with no visible measurement of arrival; only at the edge of reach does the long dwell keep buying wins.

\section{Changing the evidence}
\label{sec:intervention}

\begin{figure*}[t]\centering
\includegraphics[width=5.5in]{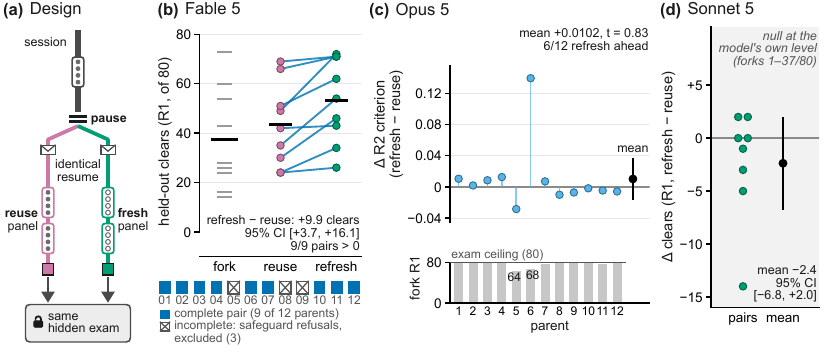}
\caption{\textbf{The panel-refresh intervention.} \textbf{(a)} Design: a session pauses at a preset point; two branches resume identically and differ only in the validation panel the harness serves, reused or fresh; both shipped programs take the same hidden exam (conditions in text). \textbf{(b)} Fable~5: refresh wins 9/9 completed pairs. \textbf{(c)} Opus~5: 12/12 pairs, null on the R2 utility (forks at the R1 ceiling); fork clears in the strip below. \textbf{(d)} Sonnet~5: null at the model's own level. Panels (b) and (d) print clears; 1 clear = 1.25 points.}
\label{fig:intervention}
\end{figure*}

\textbf{Changing only the evidence changes the product.} If the evidence an agent gathers is part of what it ships, refreshing the evidence should change the product; the panel-refresh intervention tests exactly that (Figure~\ref{fig:intervention}a). The harness pauses a session at a preset point; two branches resume from an identical message and differ only in which validation panel the harness serves: the one the parent had been reusing, or a fresh panel from the same distribution. The arm is invisible to the agent, branches run to their own end under equal caps, and both shipped programs take the same hidden exam.

In Fable~5, whose forks had headroom (mean 37.6/80), refreshed validation won all nine completed pairs of twelve initiated (Figure~\ref{fig:intervention}b), mean $+12.4$ points (95\% CI $+4.6$ to $+20.1$; attempt ledger in Appendix~L). The replication in Opus~5 ran all twelve pairs and returned a null on the pre-registered R2 endpoint ($+0.0102$, $t{=}0.83$; Figure~\ref{fig:intervention}c). Its forks sat at the R1 ceiling (median 80/80); descriptively, the two pairs forked well below that ceiling (64 and 68 of 80) moved most on R2, one down and one up ($-0.028$, $+0.139$). Sonnet~5's eight pairs bracket zero (CI $-8.5$ to $+2.5$ points): they forked at 1--37 of 80, the model's own level, and validator uptake was low (three branches per arm never called it; Appendix~L). The supported claim is narrow: where branches had both headroom and uptake, refreshing what the final checks see improved what shipped; ceiling and uptake remain candidate explanations for the other strata (Appendix~L).

\section{Related work}
\label{sec:related}

\textbf{Agents doing research.} The closest lane studies agents as autonomous researchers of machine learning itself \citep{lu2024aiscientist,hambardzumyan2026aira2,askarbekuly2026autoresearch,huang2024mlagentbench,chan2025mlebench,nathani2025mlgym,starace2025paperbench}. We move the object of research from open-ended ML to closed instrumented worlds where frozen artifacts are re-scored on hidden exams and the agent's evidence is checked line by line.

\textbf{Agents building artifacts.} Harness- and artifact-development benchmarks score what agents build \citep{wu2026harnessdev,zheng2026seagym,ursekar2026harnessopt,wang2026rethinkingharness,lu2026metaagent,chi2026gamedevbench}, and game-playing work scores how models play: directly \citep{paglieri2025balrog,hu2026lmgame,ying2026gamestore,guertler2025textarena,kuttler2020nle}, through scaffolds \citep{wang2023voyager,tan2025cradle}, and in full strategy games \citep{vinyals2019alphastar,sun2018tstarbots,ma2024textsc2,qi2024civrealm,chen2025voxdeorum,bakhtin2022cicero}; \citet{hu2024gameagentsurvey} survey; full sweep in Appendix~N. Gauntlet sits between: the deliverable is a program that plays model-free, as in programmatic policy work \citep{verma2018pirl,xu2025portal,deng2024smacr1,kuang2025gamecodeopt} and evolutionary program search \citep{romeraparedes2024funsearch,novikov2025alphaevolve}, but the object measured is the session that produced it. The nearest neighbors also study the building process: EvoPolicyGym evolves complete Python policies under a fixed 128-episode budget with platform-served feedback and best-validation checkpointing \citep{wang2026evopolicygym}; CodeClash scores agent-built codebases over fixed tournament rounds \citep{yang2026codeclash}; chess-engine builders show optimistic self-assessment against independent Elo \citep{acher2026chess}; GPT-5.4 Connect-4 builders under-use their allotted time \citep{sherwood2026connectfour}. Gauntlet adds uninstructed sessions in which the agent supplies its own validation evidence, independent replay of frozen versions on hidden exams, and an intervention on that evidence itself.

\textbf{Reusing evidence and extrapolating curves.} Our validation findings instantiate, in the wild, what the adaptive-data-analysis literature predicts for reused holdouts \citep{dwork2015reusable,blum2015ladder,cawley2010overfitting} and what agent studies report as optimistic self-evaluation \citep{park2026stagnation}; our curve and forecasting analyses adapt learning-curve extrapolation \citep{domhan2015extrapolation,swersky2014freezethaw,rakotoarison2024icft} and test-time-scaling curves \citep{liu2026elopertoken} to within-session development. Agent stopping has been studied as premature abandonment or runaway looping \citep{chen2026vigil,hou2026infiniteloops,sogani2026webagentsfinish}; the transcript-budget regularity and its race with evidence appear to be new. Normative stopping rules, marginal value \citep{charnov1976marginal} and optimal search \citep{weitzman1979search}, supply the background these measurements can now be set against. Compaction's behavioral effects have been noted \citep{tamba2026compaction}; we measure the trigger and what crossing it does. (GauntletBench, a GUI-task benchmark, is unrelated \citep{vysotskyi2026gauntletbench}.)

\section{Discussion and limitations}
\label{sec:discussion}

Read as a researcher, the coding agent has recognizable pathologies: it measures honestly, then writes an optimistic abstract; it validates on instances it already knows; its final battery is a ritual of confirmation; and it stops when its transcript budget runs out far more often than when the evidence says done. Each pathology suggests an instrumented remedy; one, serving fresh validation evidence, is shown here to change what ships. The effective transcript budget should interest harness designers: model-specific, stable across the games we can test, and uninstructed, it predicts the end of hard sessions better than any visible evidence signal; whether it can be steered is testable. Compiled agency is what these sessions deliver; the experiments the agent designs are part of it, and Gauntlet measures that part.

The limits: exam ceilings compress the top of the leaderboard, and one replication's null came with forks at that ceiling; the open questions move to Cheater:Insane and Civilization. Cells are small ($N{=}1$--$10$; frontier 6); every number is reported per cell. The race model is post hoc (the registered test is the hazard analysis). Agent-view recovery is 86\%, and frontier reasoning is mostly redacted: trace audits read behavior, not thought. One intervention in one game and one model with headroom is evidence of a mechanism, not a law; seed reuse stales evidence only because exams are deterministic. Finally, a harness contribution to cross-vendor differences cannot be excluded (\sect\ref{sec:setup}).

\bibliographystyle{iclr2027_conference}
\bibliography{refs_landscape,refs_additions}

\end{document}